\documentclass[10pt,a4paper,footsepline,headsepline,headinclude,twocolumn]{scrartcl}

\usepackage{setspace}

\usepackage{amsmath,amssymb,amsthm,amsfonts,mathrsfs}
\usepackage{a4wide}
\usepackage{scrpage2}

\usepackage[english]{babel}
\usepackage[T1]{fontenc}
\usepackage[utf8x]{inputenc}

\setfootsepline{.5pt}

\usepackage{graphicx}
\usepackage{color}

\graphicspath{{./}} 
\usepackage[hangindent=0pt,singlelinecheck=on,format=plain,font=scriptsize]{caption}
\usepackage[font=scriptsize,labelformat=parens]{subcaption}

\usepackage{float}
\usepackage{wrapfig}
\restylefloat{figure}

\usepackage[algoruled,linesnumbered]{algorithm2e}
\makeatletter
\newcommand{\removelatexerror}{\let\@latex@error\@gobble}
\makeatother

\newcommand{\SDalgorithm}{
\begingroup
\removelatexerror
\begin{algorithm*}[H]
  \label{alg-SD}
  \setstretch{1}
  \RestyleAlgo{ruled}
  \SetAlgoLined
  \LinesNumbered
  \SetKwInOut{Input}{input}
  \SetKwInOut{Output}{output}

  \Input{Shapes $\mathcal{S},\mathcal{T}$, tolerance $\eps$, elasticity
    parameters.}

  \Output{Deformation field $\Phi = \id + u$, deformed meshes
    $\Phi(\mathcal{B}^h)$ and boundary forces $f_B$.}

  \Begin{

    Build a triangulation $\mathcal{B}^h$ from the shape outline
    $\mathcal{S}$\;

    Compute the stiffness matrix of the linear material. Then compute
    the linear relation between boundary displacements and boundary forces
    according to~(\ref{eq:5-4c})\;

    Initialize $k=0$ and $u_k^h=0$\;

    Choose $\alpha$ and $\beta$ initially\;

    \While{$k\leq K_{\text{max}}$ and $\mu((\mathcal{S}+u_k^h(\mathcal{S}))\sd
      \mathcal{T})>\eps$}{
    
      Compute the measure of the symmetric difference
      $\mu((\mathcal{S}+u_k^h(\mathcal{S}))\sd \mathcal{T})$ and its gradient
      at $u_k^h$ according to Section~\ref{s-opt-compute}\;
    
      Find the optimal displacement field $u_{k+1}^h$ that solves
      problem~(\ref{eq:5-7}) (with $u_0^h=u_k^h$)\;
    
      Increase $\alpha \leftarrow q\alpha$ and $\beta \leftarrow q\beta$ by a
      pre-selected constant $q\approx 1.2$ to $1.4$\;
    
      Set $k\leftarrow k+1$\;
    }
   
    Compute the full deformation of the source shape $\Phi(\mathcal{B}^h)$
    by computing the interior displacements according to~(\ref{eq:5-10})\;

    Return\;
  }
  
  \caption{Sketch of the symmetric difference algorithm}
\end{algorithm*}
\endgroup}

\usepackage[colorlinks=true]{hyperref}

\usepackage{dblfloatfix}
\usepackage{fixltx2e}

\usepackage{lipsum} 
\usepackage{authblk}

\theoremstyle{plain}

\theoremstyle{definition}

\theoremstyle{remark}

\newcommand {\real}{\mathbb{R}}

\newcommand {\id}{\mathbb{I}}
\newcommand {\sd}{\bigtriangleup}

\newcommand {\eps} {\varepsilon}

\DeclareMathOperator*{\minimize}{min}

\DeclareMathOperator{\DIV}{div}

\newcommand{\tr}[1]{\operatorname{tr}#1} 

\newcommand{\abs}[1]{\left\lvert #1 \right\rvert}

\newcommand{\dual}[2]{\left\langle #1, #2\right\rangle}
\newcommand{\norm}[2]{\left\| #1 \right\|_{#2}}

\renewcommand{\d}{\:\mathrm{d}}

\lohead{}
\rohead{Symmetric Difference Minimization for Shape Matching}
\rofoot{\pagemark} \lofoot{} 

\begin{document}

\thispagestyle{empty}

\vspace{1.0cm}

\begin{center}
  \LARGE \bf{Elasticity-based Matching by Minimizing the Symmetric Difference
    of Shapes}
\end{center}
 
\vspace*{1cm}

\begin{center}
  Konrad Simon\textsuperscript{$\ast$} and Ronen Basri\textsuperscript{$\ast$}
\end{center}

\vskip 1cm  

\begin{center}
  \small {	\textsuperscript{$\ast$}~Department of Computer Science \\ and Applied Mathematics, \\
  	The Weizmann Institute of Science, Rehovot 76100, Israel \\[0.5cm]
	}
\end{center}

\vspace{1cm}

\thispagestyle{empty}

\begin{abstract}
  \textbf{Abstract.} \small We consider the problem of matching two shapes
  assuming these shapes are related by an elastic deformation. Using
  linearized elasticity theory and the finite element method we seek an
  elastic deformation that is caused by simple external boundary forces and
  accounts for the difference between the two shapes. Our main contribution is
  in proposing a cost function and an optimization procedure to minimize the
  symmetric difference between the deformed and the target shapes as an
  alternative to point matches that guide the matching in other techniques. We
  show how to approximate the nonlinear optimization problem by a sequence of
  convex problems. We demonstrate the utility of our method in experiments and
  compare it to an ICP-like matching algorithm.
\end{abstract}

\vspace{0.7cm}


\setcounter{page}{1}

\section{Introduction}\label{s-1}

\subsection{Motivation}\label{s-1-1}

Understanding two-dimensional shape and its variations is one of the
fundamental problems in computer vision, computer graphics, and medical
imaging. In particular, finding correspondences between two given input shapes
is the basis of many applications such as recognition and information
transfer. One way for finding correspondences between shapes is to compute a
deformation that aligns one shape (the source) with the other (the
target). This problem is usually referred to as the shape matching problem.

From a mathematical stand point the shape matching problem is an ill-posed
inverse problem. This is because a desired matching is often tied with the
unknown semantics of the given shapes. The usual way to cope with this is
though regularization, i.e., one balances the quality of a correspondence
against its regularity.


\subsection{Our Method}\label{s-1-2}

In this work we propose an alternative shape matching algorithm in two
dimensions to ICP and ICP-like methods. In addition we will provide a
comparison to our ICP-like methods, recently introduced
in~\cite{Simon,Simon2}. Our approach minimizes the area of mutual non-overlap,
i.e., the area of the symmetric difference of the compared shapes. This is a
global dissimilarity measure and a low value, in contrast to ICP, assures a
good (but not necessarily meaningful) alignment of the deformable source and
the fixed target shape.

To find meaningful alignments we use linearized elasticity as a regularizer
for the deformation. More specifically, we seek to explain deformations of the
source shape by means of elastic forces acting on the shape boundary. This is
a reasonable strategy since many shapes in applications depict actual physical
bodies and hence shape change can be explained by means of external causes. We
furthermore believe that in many scenarios an observed deformation can be
explained by very simple causes, i.e., the physical forces that cause the
shape change are simple although the deformation itself may not be simple. The
reader may think of articulations of a given object, such as a human shape or
a moving animal, where forces mostly act on the articulated parts. This
motivates to look for boundary forces that are sparse while simultaneously
seeking a good alignment. This is in the spirit of~\cite{Sederberg}, as one
usually seeks a good alignment by means of minimal cost.

The optimization problem we pose is nonlinear. We show a heuristic way to
approximate the problem by a sequence of (convex) second order cone problems
(SOCPs). Our shapes are represented by their outlines which we use to build
triangular meshes of them. The finite element method (FEM) then allows us to
discretize the underlying equations of linearized elasticity, the
Navier-Lam\'e equations, and to derive a linear relation between (nodal)
boundary displacements and (nodal) boundary forces. This is done by a
reduction of the stiffness matrix and yields a suitable representation of the
elastic regularizer that only involves the boundary.

The area of the symmetric difference is computed by first finding its
polygonal outline and then computing the area by means of the divergence
theorem. The polygon describing the symmetric difference is computed using
Vatti's clipping algorithm~\cite{Vatti}, which was originally used in the
context of rendering. Note that the triangular mesh is only needed for the
initial step of the FEM, i.e., the computation of the stiffness matrix. The
optimization, though, takes place on the boundary (of the mesh) only, which
reduces the number of variables while keeping information of the elastic
properties of the interior of the source shape.

\subsection{Related Work}\label{s-1-3}

\paragraph{\bf Overview.}
An optimization based shape matching algorithm is, as already insinuated,
usually based on the principle of seeking a good alignment while keeping a
deformation cost, measured on an admissible set of deformations, as low as
possible.

Rigid matching, i.e., the admissible deformations are translations and
rotations, is considered to be
well-understood~\cite{Aiger,Gelfand,Rusinkiewicz}. These low-dimensional
deformations are the first choice when it comes to the alignment of range
scans of rigid objects.

In the case of non-rigid matching there is a larger variety of deformation
models. The authors of~\cite{Beg,Christensen}, for instance, model a
deformable medium as a fluid that gradually deforms into the desired
target. This was done in the context of medical image matching and has the
advantage that it allows very large deformations. Elastic regularizers, on the
other hand, are in clear contrast to fluid regularizers as they do carry
information about previous deformation states (fluids do not have
``memory''). They have been pioneered in computer graphics by
Terzopoulos~\cite{Terzopoulos} and in medical imaging by Broit and
Bajcsy~\cite{BajcsyBroit,Gee}. In particular, in the context of matching,
elasticity based (or related) methods have been used
in~\cite{Amit1,BajcsyBroit,Ferrant,PeckarSchnorrRohrStiehl}. Overviews can be
found in~\cite{Holden,Nealen}.

Both, fluid and elasticity methods model shape as a continuum and hence
contain information of the behavior of their interior. Shapes can also be
outlined by their contours such as polygonal lines. Matching then amounts to
curve matching and is usually a lower dimensional problem due to the compact
curve representation. Younes~\cite{Younes1,Younes2} models curves by their
associated angle functions and a group action on this representation models
deformation. An energy modeling elastic behavior of the curve is then
minimized to obtain correspondences. The authors of~\cite{Mio} model a
Riemannian structure on a manifold of curves and match them by finding
geodesics on this manifold. The geodesic distance is then used as a comparison
measure between shapes. Note that modeling shapes as curve outlines does not a
priori take into account their interior. A good survey on curve matching can
be found in~\cite{Younes3}.

\vspace{0.5cm}

Measuring the alignment of the shapes is the second ingredient for an
optimization based algorithm. A well-known method is the so-called iterative
closest point algorithm introduced by Besl and McKay~\cite{Besl}. ICP is an
iterative method that aligns shapes by altering between finding a good
transform and updating nearest neighbor correspondences. This amounts to
minimizing mean distance of the shapes by means of nearest neighbor
correspondences. It has been used originally for rigid matching. Refined
variants based on different correspondence rejection and weighting methods can
be found in~\cite{Rusinkiewicz}. For more recent non-rigid variants of ICP
see~\cite{Allen,Brown,Kovalsky}. Our methods introduced in~\cite{Simon,Simon2}
are modifications of classical ICP since we allow estimated correspondences to
drift during the optimization instead of keeping them fixed. Therefore, we
refer to them as ICP-like methods.

The nearest neighbor distance used in ICP can be seen as the $L^2$-version of
the one-sided Hausdorff distance~\cite{Dubuisson}. It is not symmetric and
hence not a metric. In contrast, the area of symmetric difference constitutes
a metric on the space of shapes. Minimizing a dissimilarity metric for
matching is called metric matching. A theoretical result for matching convex
shapes measured by the symmetric difference can be found in~\cite{Alt}. There
the authors show that aligning the centroids of the shapes guarantees that
their symmetric difference is within a certain bound of the
optimum. Unfortunately this only holds for a rather restrictive set of
transforms and shapes. More practically oriented approaches can be found in
the context of segmentation and registration using level set
methods~\cite{Sethian}, see~\cite{ChanVese1,ChanVese2,Overgaard}. There, the
shapes are represented as level set functions, and functionals derived from
the Chan-Vese model~\cite{ChanVese1,ChanVese2} are minimized. The area/volume
of the symmetric difference was also used for finding a shape
median~\cite{Berkels} and in the context of elastic shape
averaging~\cite{Rumpf1}. These approaches are (also) closely related to
minimizing the Mumford-Shah functional~\cite{Mumford}. In~\cite{Berkels} a
Chan-Vese model is employed and in~\cite{Rumpf1} shapes are encoded by means
of a so-called phase field function~\cite{Ambrosio}.

Other metric dissimilarity measures in the literature, used for metric
matching, include the Hausdorff distance~\cite{Huttenlocher}, the
Gromov-Hausdorff distance~\cite{Bronstein2,Memoli} and the Gromov-Wasserstein
distance\cite{Memoli,Memoli2}. These metrics are either rather hard to compute
or can suffer from robustness issues.

\paragraph{\bf Comparison.}
The work presented in this paper borrows methods from linearized elasticity to
measure the difficulty of shape change. The splitting of the stiffness matrix
that we employ to minimize boundary forces was previously used in the context
of surgery simulations~\cite{Bro-Nielsen} and in our approach to
two-dimensional shape matching in~\cite{Simon}. In contrast to minimizing the
elastic energy or to computing geodesic lengths, which is done in most of the
above cited works that employ elasticity, seeking unknown forces has the
advantage that we get an explanation of the unknown deformation in addition to
the deformation itself. In particular sparse forces, we believe, can account
for simple explanations. This principle was generalized in~\cite{Simon2} in
the ICP-like setting to nonlinear elastic matching in three dimensions.

Both of our previous ICP-like schemes employ (descriptor aided) nearest
neighbor correspondences to drive the search for an alignment. The difference
to our symmetric difference method is that for the latter we do not use any
kind of ``guessed'' correspondences or other shape information like curvature
or descriptors. This renders the method simple. Both methods have in common
that the estimated correspondences and the negative gradient of the area of
the symmetric difference, which we use in the optimization, can be interpreted
as a restoring force between source and target that needs to be
regularized. Furthermore, our previous methods and our symmetric difference
method can be classified as local methods since they converge to a local
minimum that depends on the initial alignment of the shapes.

We provide experimental evidence that our method can outperform the ICP-like
method~\cite{Simon} in two dimensions by means of robustness and
alignment. Indeed, the ICP-like method is not guaranteed to give a good
alignment since it can theoretically converge to a single point on the target
shape. In contrast, a low area of the symmetric difference always means a good
(and by means of regularization hopefully meaningful) overlap.

Our symmetric difference method relates to methods used for metric
matching~\cite{Berkels,ChanVese1,ChanVese2,Overgaard,Rumpf1} but uses
completely different tools. All the above cited literature represent shapes
using a level set approach and make use of their indicator functions. This
amounts to an area based method. For the computation of the symmetric
difference and its area we, in contrast, solely make use of its polygonal
outline. For this an efficient version of Vatti's algorithm for polygonal
clipping is the essential ingredient. Since we also reduce the elastic
regularizer to the boundary of the source shape this essentially amounts to
matching closed curves. Note that on the other hand we keep elastic
information of the interior of the shape and can compute interior
displacements by simply solving a linear system of equations. This way we take
advantage of the lower dimension of the optimization problem on the boundary
while not losing information of the interior of the shape.

\vspace{0.5cm}

This work is organized as follows. In Section~\ref{s-2} we give an overview of
the relevant part of elasticity theory and of the properties of the area of
symmetric difference. We also give a detailed explanation of the optimization
problem that we solve in a heuristic manner. Section~\ref{s-3} gives an
overview of the implementation and shows results of experiments with shapes
taken from various data sets. A direct comparison to our method introduced
in~\cite{Simon} is provided. Section~\ref{s-4} concludes with a discussion.


\section{Methods}\label{s-2}

\subsection{Linearlized Elasticity and FEM}\label{s-2-elasticity}

We give a concise overview of the parts of linearized elasticity and the
finite element framework that are relevant to this work. Additional and more
detailed material about elasticity theory can be found
in~\cite{Ciarlet,Dhondt}. For more material on FEM see~\cite{Braess,Dhondt}.

\begin{figure}[h]
  \centering
  \includegraphics[width=7cm]{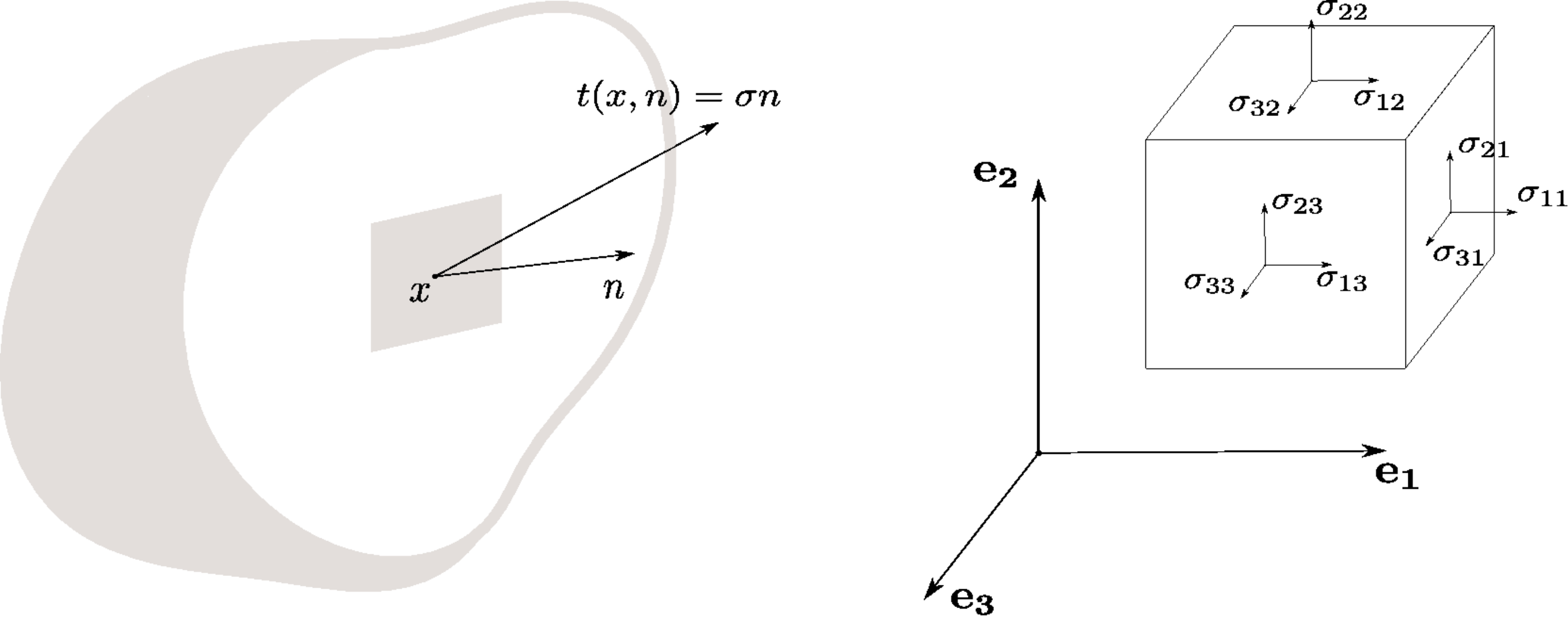}
  \caption{Left: an illustration of a surface force $t(x,n)$ at some point $x$
    in an arbitrary cross section with normal $n$ of a body $B$.  Right:
    illustration of the three components of the stress tensor with respect to
    the canonical coordinate planes. The normal stress is orthogonal to the
    considered plane shear stresses lie within the plane.}
  \label{fig-stress}
\end{figure}

We begin by explaining linearized elasticity theory in 3D. Elasticity theory
explains the states of elastic bodies that are subject to external forces. An
elastic body is a physical entity, described by a closed connected and bounded
set $\mathcal{B}\subset\real^3$, that reacts to external forces with a
deformation and returns to its original shape after the forces are removed. An
elastic body has no memory of previous deformations or applied forces.

The elasticity equations describe the balance between external forces $g$,
applied to a an elastic body, and internal forces that resist the deformation
caused by the external ones. The basic equation reads
\begin{equation}
  \label{eq:el-1}
  \begin{split}
      -\DIV \sigma & = g \quad \text{in $\mathcal{B}$} \:, \\
  \end{split}
\end{equation}
where $\sigma$ is called the Cauchy stress tensor (or simply stress),
represented by a $3$-by-$3$-matrix. Roughly speaking, $\sigma$ measures the
internal stress distribution, i.e., for any $x\in\mathcal{B}$ it yields the
resulting surface traction $t(x,n)$ (force measured per unit area) among all
cross-sections given by their surface unit normal $n$. The stress is symmetric
and material specific. An illustration of the traction is given in
Figure~\ref{fig-stress}.

The material specific reaction of a body to applied forces in linearized
elasticity is modeled by a linear relation between strain and stress. Strain
measures the local change of lengths inside a solid body. It is rotation
invariant and can hence not be linear. In linearized elasticity we use its
first order approximation which is given by $\eps=1/2(\nabla u + \nabla u^T)$
where $u:\mathcal{B}\rightarrow\real^3$ is the displacement field, i.e.,
$x\in\mathcal{B}$ moves to $\Phi(x)=x+u(x)$. $\Phi$ is the so-called
deformation field. The strain measure $\eps$ is not invariant under general
Euclidean motions but only under translations and infinitesimal rotations. The
linear relation between stress and strain is usually encoded in a fourth-order
tensor but assuming homogeneity and isotropy of the solid it can be simplified
to Hooke's law:
\begin{equation}
  \label{eq:el-2}
  \sigma = \lambda\tr{(\eps)}\id + 2\mu\eps \:.
\end{equation}
The constants $\lambda$ and $\mu$ are the Lam\'e constants, $\tr{(\cdot)}$ is
the trace and $\id$ denotes the identity. This model is often used in
engineering since it approximates the elastic behavior of solids very well
provided the deformation is small. 

Combining the expressions for $\eps$ and Hookes's law with~(\ref{eq:el-1}) we
get the Navier-Lam\'e equation
\begin{equation}
  \label{eq:el-3}
  \mu \Delta u + (\lambda + \mu)\nabla\DIV u + g = 0 \quad \text{in $\mathcal{B}$}
\end{equation}
supplemented with the boundary conditions
\begin{equation}
  \label{eq:el-4}
  \begin{split}
    u & = u_0 \quad \text{on $\Gamma_D$,} \\
    \sigma(u)n & = f \quad\;\: \text{on $\Gamma_N$} \\
  \end{split}
\end{equation}
where $\Gamma_D$ and $\Gamma_N$ are disjoint portions of the boundary
$\partial\mathcal{B}$, $n$ is the outward unit normal on $\Gamma_N$, and $f$
is a known external surface force. This PDE constitutes a linear elliptic
system and is well-posed~\cite{Braess} (up to translation and infinitesimal
rotation if $\Gamma_D=\emptyset$). For our case of planar shape matching we
simply set the third component of the displacement field to zero, i.e., we set
$u=(u_1,u_2,0)$. This leaves the structure of~(\ref{eq:el-3}) unchanged.

For a numerical solution of equation~(\ref{eq:el-3}) on general domains we use
a piece-wise continuous finite elements on a triangular mesh. Conformal finite
elements rely on the weak form of the underlying PDE which is given in our
case as: find $u$ that satisfies the boundary conditions~(\ref{eq:el-4}) on
$\Gamma_D$ such that
\begin{equation}
  \label{eq:el-5}
  \begin{split}
    \int_{\mathcal{B}} & \mu\dual{\nabla u}{\nabla v} + (\lambda+\mu)(\DIV
    u)(\DIV v)
    \d x \\
    & = \int_{\Gamma_N}\dual{f}{v} \d S - \int_{\mathcal{B}}\dual{g}{v} \d x \quad
    \forall v \in V
  \end{split}
\end{equation}
where $V$ is a space of suitable test functions. For details
see~\cite{Braess,Dhondt}. Conformal FEMs replace $V$ with a finite dimensional
approximation $V^h$. Choosing a basis in $V^h$ one can
transform~(\ref{eq:el-5}) into a linear system of equations for the nodal
displacements $u^h$
\begin{equation}
  \label{eq:el-6}
  Au^h = f^h - g^h \:.
\end{equation}
Note, that the traction boundary conditions enter the right-hand side. The
system matrix $A$ is called stiffness matrix and is symmetric positive
semi-definite. Its rank in case of planar elasticity is $2N-3$ where $N$ is
the number of nodes in the mesh and its kernel consists of translations and
infinitesimal rotations.

\vspace{0.5cm}

\subsection{Properties of the Symmetric Difference}\label{s-2-symDiff}

As a comparison measure between shapes the symmetric difference has a few
attractive properties. Most importantly, it constitutes a metric on the shape
space. This can easily be seen as follows: the symmetric difference as an
operation on the power set $\mathcal{P}(X)$ of any given set $X$ makes it an
Abelian group with the empty set as the neutral element. Any set
$A\in\mathcal{P}(X)$ is inverse to itself. The map $A\mapsto\chi_{A}$, where
the right-hand side denotes the indicator function of $A$, defines a group
homomorphism from $\mathcal{P}(X)$ into the group of indicator functions
endowed with the binary relation $\chi_A\sd\chi_{B}:=
\abs{\chi_A-\chi_B}$. Restricting this homomorphism to the set of measurable
functions one can express the Lebesgue measure of the symmetric difference as
\begin{equation}
  \label{eq:5-1}
  \mu(A\sd B) = \int_{\real^d} \abs{\chi_A - \chi_B} \d x = \int_{\real_d}
  \chi_{A\sd B} \d x \:.
\end{equation}
This way it is easy to see that the measure of the symmetric difference fulfills the
triangle inequality
\begin{equation}
  \label{eq:5-2}
  \mu(A\sd B) \leq \mu(A\sd C) + \mu(C\sd B)
\end{equation}
and hence $\mu(\cdot\sd \cdot)$ is a pseudo-metric on the set of measurable
functions. Identifying all sets that are equal almost everywhere we even get a
metric.

Suppose that we are given a pair of semantically similar shapes $\mathcal{S}$
(source) and $\mathcal{T}$ (target). The goal is to find a meaningful
transformation $\Phi:\mathcal{S}\rightarrow\mathcal{T}$, i.e., a smooth and
locally injective vector field. The admissible shapes are outlined by closed
non-self-intersecting curves in two dimensions or orientable manifolds without
boundary. Both define a proper area or volume and they can be regarded as the
boundaries of solids. We denote with $\mathcal{B}$ the solid represented by
$\mathcal{S}$. In particular we have
$\partial\mathcal{B}=\mathcal{S}$. $\Phi:\mathcal{B}\rightarrow\real^3$
denotes the volumetric deformation of $\mathcal{B}$.


\section{The Optimization Problem}\label{s-opt}

\subsection{Minimizing the Area of the Symmetric Difference}

This paper focuses on the case of planar matching and further assumes zero
volumetric forces $g=0$ in~(\ref{eq:el-3}), i.e., the unknown deformation is
caused by boundary forces only. The optimization problem to be solved in our
framework can now be formulated as
\begin{equation}
  \label{eq:5-3}
  \begin{split}
    \minimize_{\Phi} & \quad \int_{\mathcal{S}} \norm{\sigma(u)n}{2} \d S \\
    \text{subject to} & \quad \mu(\Phi(\mathcal{S})\sd \mathcal{T}) = 0 \\
    & \quad \mu \Delta u + (\lambda+\mu)\nabla\DIV u = 0 \quad \text{in $\mathcal{B}$} \\
    \quad & \quad \Phi = \id + u \:.
  \end{split}
\end{equation}
The expression $\mu(\Phi(\mathcal{S})\sd \mathcal{T})$ refers to the area of
the symmetric difference that is defined by the interiors of
$\Phi(\mathcal{S})$ and $\mathcal{T}$. This optimization problem has a convex
objective, but it is nonlinear in $\Phi$ due to the constraints. We propose a
way to tackle this problem heuristically and show how to approximate it by a
sequence of convex second order cone problems.

A relaxed formulation can be written as
\begin{equation}
  \label{eq:5-4}
  \begin{split}
    \minimize_{\Phi} & \quad \int_\mathcal{S}
    \norm{\sigma(u)n}{2} \d S + \alpha
    \mu(\Phi(\mathcal{S})\sd \mathcal{T})^2 \\
    \text{subject to} & \quad \mu \Delta u + (\lambda+\mu)\nabla\DIV u = 0
    \quad \text{in $\mathcal{B}$} \\
    \quad & \quad \Phi = \id + u \:.
  \end{split}
\end{equation}

After FEM discretization the PDE constraint becomes a linear system of
equations~(\ref{eq:el-6}). Renumbering the nodes one can assume the structure
\begin{equation}\label{eq:5-4a}
  \left(
    \begin{array}{cc}
      A_{BB} & A_{BI} \\
      A_{IB} & A_{II} \\
    \end{array}
  \right)
  \left(
    \begin{array}{c}
      u_B \\ u_I \\
    \end{array}
  \right)
  =
  \left(
    \begin{array}{c}
      f_B \\ 0 \\
    \end{array}
  \right) \:,
\end{equation}
where $u_B$ and $f_B$ are respectively the vectors of the nodal displacements
and forces at the $K$ boundary nodes and $u_I$ denotes the displacements at
$N-K$ inner nodes. Note that $f_I=0$ on the right-hand side of~(\ref{eq:5-4a})
since we assumed only boundary forces. We assume the displacements
$u_i\in\real^2$ and the forces $f_i\in\real^2$ at each boundary node in
$u_B\in\real^{2K}$ and $f_B\in\real^{2K}$ to be arranged in the following
manner:
\begin{equation} \label{eq:5-4b}
  u_B = 
  \left(
    \begin{array}{c}
      u_1^{(1)} \\ u_1^{(2)} \\ \vdots \\ u_K^{(1)} \\ u_K^{(2)} \\
    \end{array}
  \right)
  \hspace{0.2cm} \text{and} \hspace{0.2cm}
  f_B =         
  \left(
    \begin{array}{c}
      f_1^{(1)} \\ f_1^{(2)} \\ \vdots \\ f_K^{(1)} \\ f_K^{(2)} \\
    \end{array}
  \right)
\end{equation}
where $u_i^{(j)}$ and $f_i^{(j)}$ denote the $j$-th component of the
displacement (force resp.) at boundary node $i$. By taking the Schur
complement with respect to the boundary block $A_{BB}$ we get the relation
\begin{equation}\label{eq:5-4c}
  Su_B = f_B
\end{equation}
where
\begin{equation}\label{eq:5-4d}
 S = A_{BB} - A_{BI}A_{II}^{-1}A_{IB} \:.
\end{equation}

We thus completely eliminated the linear system as a constraint and reduced
the problem to the boundary curve $\mathcal{S}=\partial\mathcal{B}^h$
only. This idea was previously used in~\cite{Bro-Nielsen,Simon}. The elastic
regularizer is hence reduced to the boundary while keeping information of the
elastic properties of the interior. The interior displacements can be found by
solving a linear system.  In the discretized setting the boundaries
$\mathcal{S}$ and $\mathcal{T}$ are simply polygonal lines. Taken together,
the optimization problem takes the form
\begin{equation}
  \label{eq:5-5}
  \begin{split}
    \minimize_{u^h} \quad & \sum_{i=1}^K \norm{S_iu_B}{2} \d S \\
    & \quad + \alpha\mu((\mathcal{S}+u^h(\mathcal{S}))\sd \mathcal{T})^2 \:.
  \end{split}
\end{equation}
The first term of the objective~(\ref{eq:5-5}) is already in a convenient form
but the second one is nonlinear in $u^h$ which represents the interpolated
nodal boundary displacements $u_B$ acting on $\mathcal{S}$. We replace the
second term in the objective by its first-order approximation around a
given displacement field $u_0^h$ on the boundary of $\mathcal{S}$:
\begin{equation}
  \label{eq:5-6}
  \begin{split}
    \minimize_{u^h} \quad & \sum_{i=1}^K \norm{S_iu_B}{2} \d S \\
    & + \alpha\left[ \mu((\mathcal{S}+ u_0^h(\mathcal{S}))\sd
      \mathcal{T}) \right. \\
      & + \left. \nabla\mu((\mathcal{S}+u_0^h(\mathcal{S}))\sd
      \mathcal{T})(u^h-u_0^h) \right]^2 .
  \end{split}
\end{equation}
This is similar to replacing the $L^2$-difference $\norm{I_1\circ (\id+u) -
  I_2}{}^2$ of two images $I_1,I_2$ that are to be registered by a transform
$\Phi=\id+u$ by its first-order approximation around $u=0$. This was done
in~\cite{Ferrant} in the context of unimodal medical imaging. We further
explain how we compute the symmetric difference in
Section~\ref{s-opt-compute}.

\begin{figure*}[t!]
  \centering
  \includegraphics[width=0.95\textwidth]{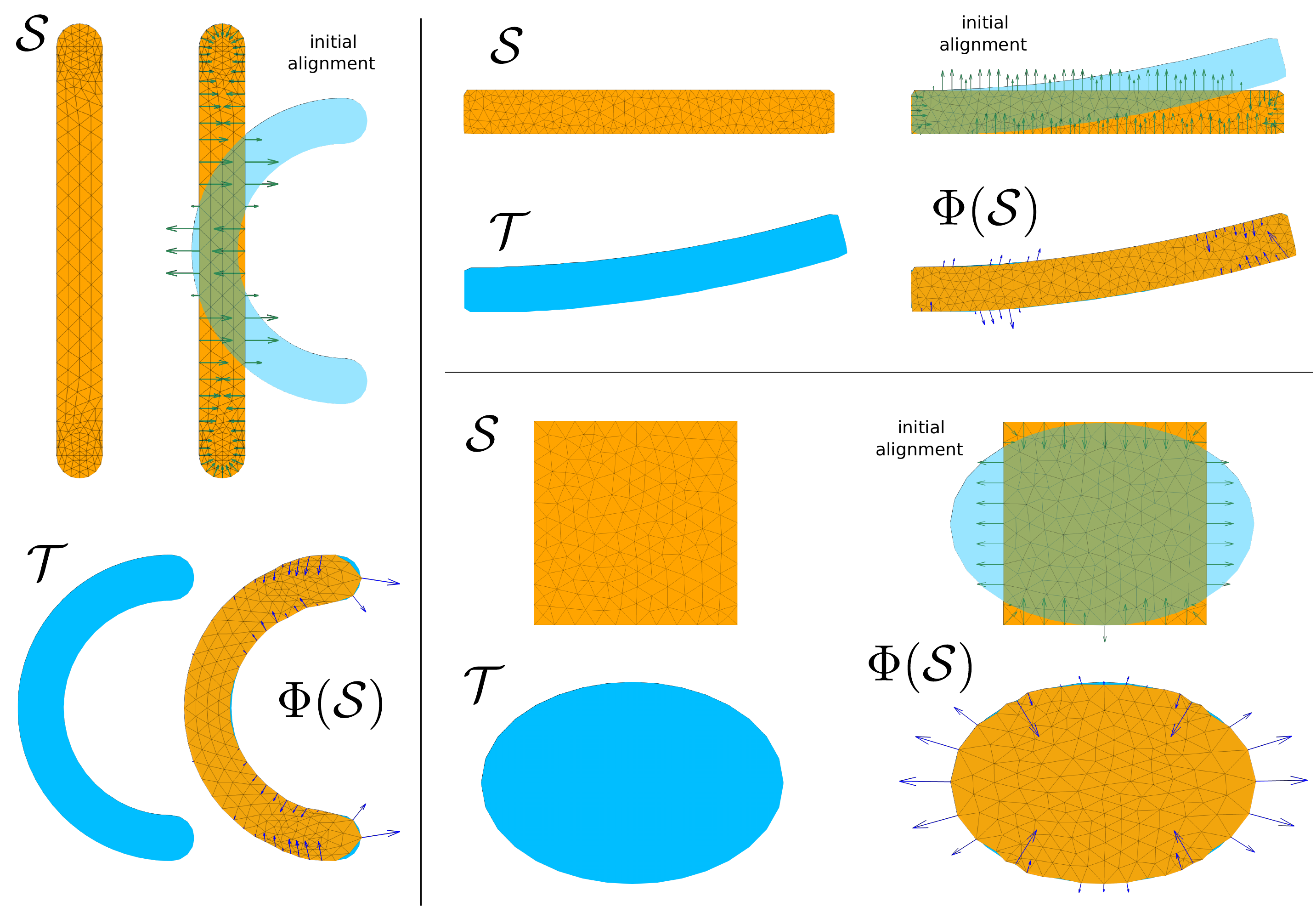}
  \caption[Symmetric difference matching: motivating examples]{We match three
    simple source shapes $\mathcal{S}$ to target shapes $\mathcal{T}$. For
    each experiment we show, in addition, the initial alignment of the shapes
    together with the negative gradient of the area of the symmetric
    difference (green arrows) as well as an overlap of the matched shape
    $\Phi(\mathcal{S})$ and the target together with the elastic forces (blue
    arrows). The beam-to-C experiment (left) and the ellipse-to-rectangle
    experiment (lower right) require relatively large deformations.}
  \label{c-4-fig-1}
\end{figure*}

\SDalgorithm
\newpage

\begin{figure}[h]
  \begin{center}
    \fbox{
      \includegraphics[width=0.45\textwidth]{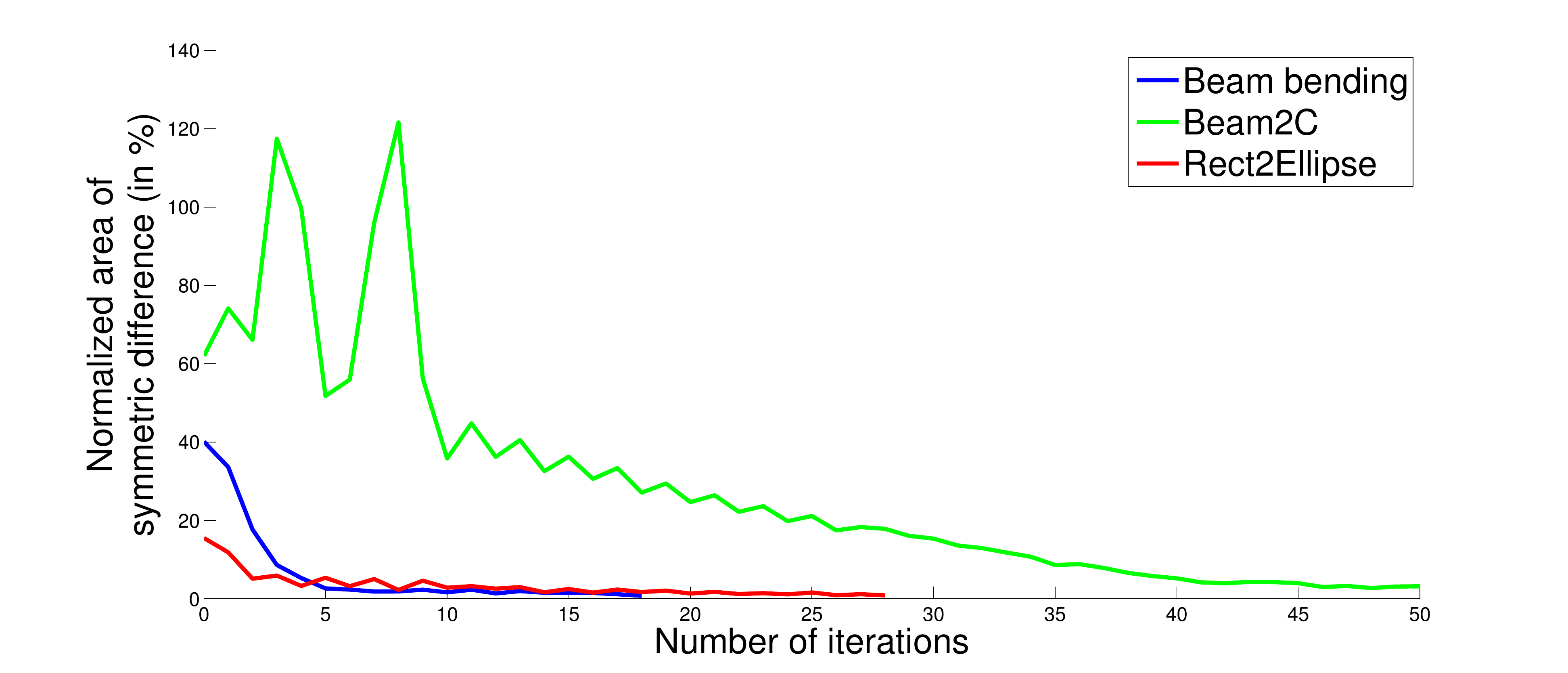}
    }
  \end{center}
  \caption[Symmetric difference matching: motivating examples (graphs)]{Area
    of the the symmetric difference in \% of the sum of the areas of the
    source and the target for the experiments shown in
    Figure~\ref{c-4-fig-1}.}
  \label{c-4-fig-1g}
\end{figure} Note that without regularization the minimum
of~(\ref{eq:5-6}) would be zero and the minimizer is not unique and can in
principle be a transform with very bad regularity properties. Furthermore, the
linear elastic regularizer we use is invariant under translations and
infinitesimal rotations. If now any of these transforms is contained in the
orthogonal complement of the space spanned by the gradient of $\mu$ at
$u_0^h$, i.e., $\nabla\mu((\mathcal{S}+u_0^h(\mathcal{S}))\sd \mathcal{T})v$
is zero or very small in magnitude, one can still find large displacements
that (almost) leave the objective~(\ref{eq:5-6}) untouched but do
significantly increase the area of the symmetric difference. But this is
exactly what we want to avoid. Therefore, we add a localization term
to~(\ref{eq:5-6}) to keep the optimal displacement field in average close to
$u_0^h$ which we used as a basis for the linearization. The problem then
becomes

\vspace{0.3cm}

\begin{equation}
  \label{eq:5-7}
  \begin{split}
    \minimize_{u^h} \quad & \sum_{i=1}^K \norm{S_iu_B}{2} \d S + \beta
    \norm{u^h-u_0^h}{2}^2 \\
    & + \alpha\left[
      \mu((\mathcal{S}+u_0^h(\mathcal{S}))\sd \mathcal{T}) \right. \\
    & + \left. \nabla\mu((\mathcal{S}+u_0^h(\mathcal{S}))\sd
      \mathcal{T})(u^h-u_0^h) \right]^2 \:.
  \end{split}
\end{equation}

Although we can not guarantee the monotonicity of the functional we observed
in experiments that on average we decrease the symmetric difference of the
mapped source and the target shape. Note that~(\ref{eq:5-7}) is a convex
problem. We summarized the optimization strategy to solve~(\ref{eq:5-3}) in
Algorithm~\ref{alg-SD}.

\begin{figure*}[t!]
  \centering
  \includegraphics[width=0.95\textwidth]{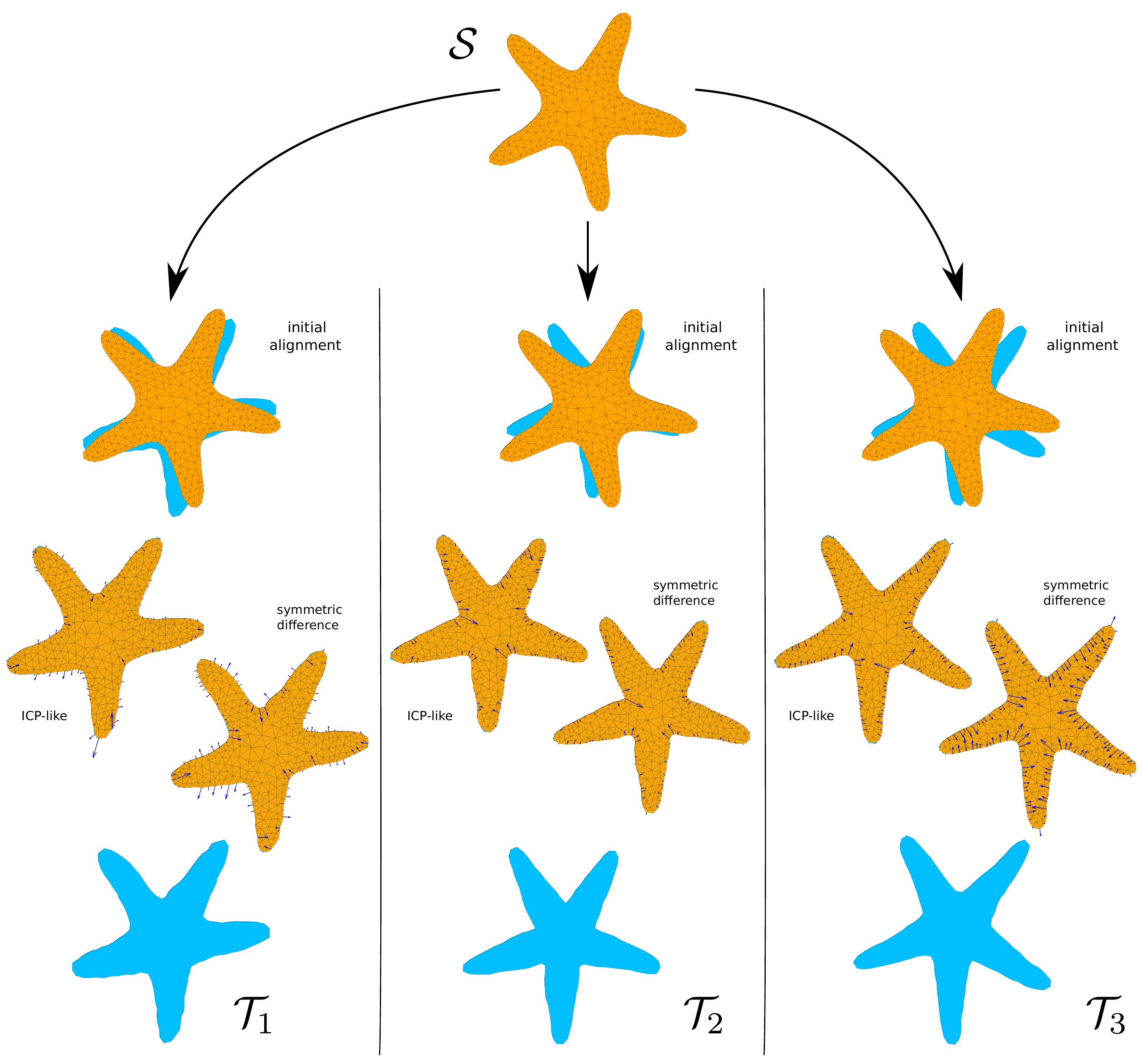}
  \caption[Starfish experiment revisited]{Three articulations of a starfish
    shape $\mathcal{S}$. We compare the results of the symmetric difference
    method to the ICP-like method. Visually the results look similar but the
    ICP-like methods exhibits higher forces and distortion. The elastic forces
    that produce the matching deformations are shown in blue together with an
    overlay of the matched shape and the target. Both iterations were stopped
    once the area of the symmetric difference dropped below $0.7$\% of the sum
    of the area of source and target.}
  \label{c-4-fig-2}
\end{figure*}

\subsection{Computing the Symmetric Difference}\label{s-opt-compute}

Looking at the objective function~(\ref{eq:5-7}) the reader can see that we
need to compute the symmetric difference of two given shapes and its
derivative with respect to a variation of one of the shapes.

We start with the area. In order to compute the exact area of the symmetric
difference one needs to know the actual symmetric difference of the two shapes
or at least their intersection. For shapes bounded by polygonal lines one can
find their intersection by means of polygonal clipping algorithms. Clipping
algorithms emerged early in computer graphics in the context of
rendering. Given a ``visible area'', outlined by a closed polygon $P_1$, and
given another polygon $P_2$, the goal \begin{figure}[h]
  \begin{center}
    \fbox{
      \includegraphics[width=0.43\textwidth]{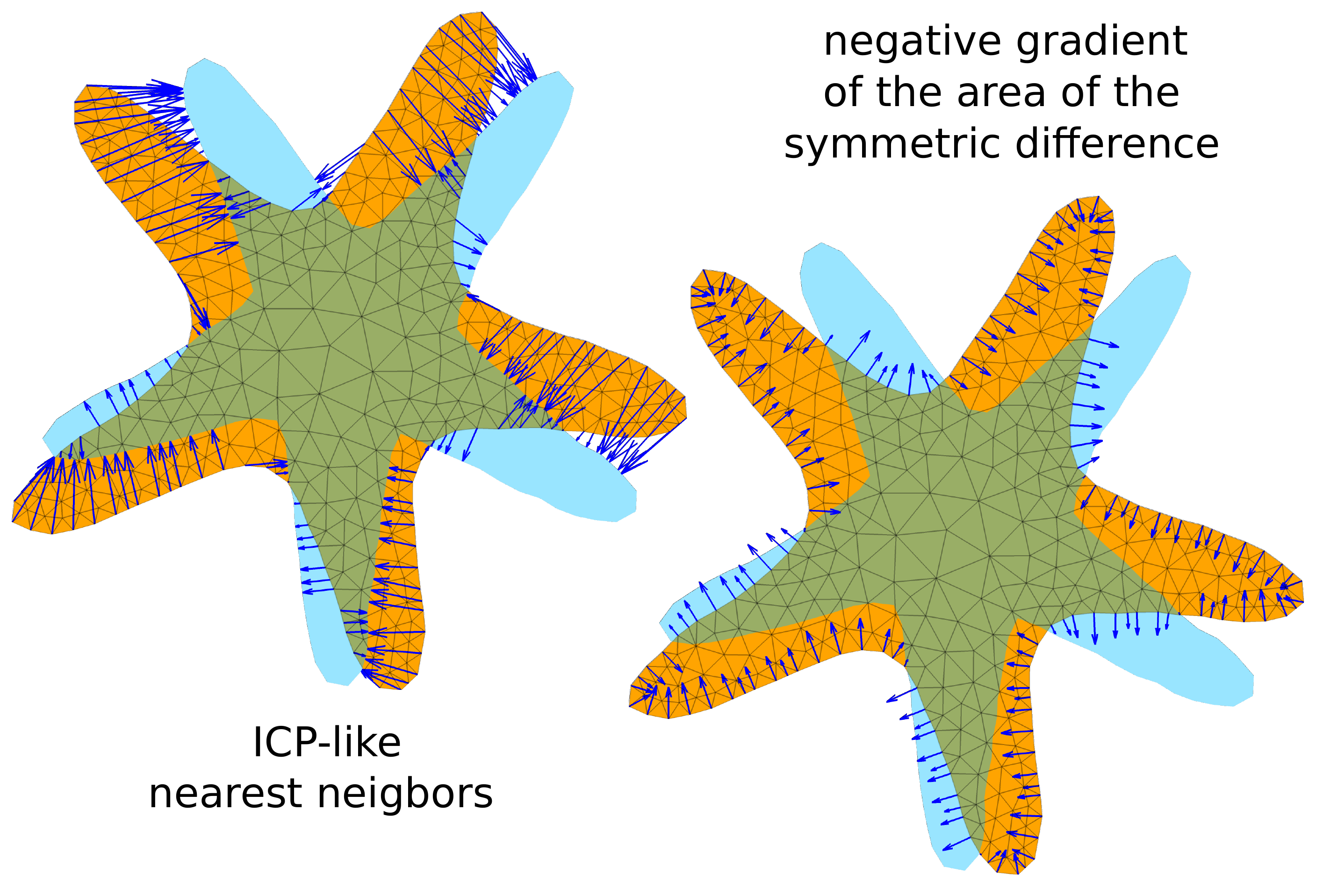}
    }
  \end{center}
  \caption[Starfish experiment: comparison of the initial restoring forces to
  our ICP-like method]{Comparison of the initial restoring forces that are
    produced by the ICP-like method and the symmetric difference method for
    target shape $\mathcal{T}_3$ in Figure~\ref{c-4-fig-2}.}
  \label{c-4-fig-2a}
\end{figure} is to remove the
``invisible'' part of $P_2$ that is outside the visible region of interest,
i.e., outside $P_1$. The visible area $P_1$ is called \emph{clip polygon} and
the polygon $P_2$ that is to be clipped is called \emph{subject polygon}. In
other words a clipping algorithm finds a new polygon $P_3$ that outlines the
area of intersection of the clip and subject polygon.

The basic idea for a convex clip polygon $P_1$ is that its interior can be
described as the intersection of a finite number of half planes described by
its edges. Hence, one can simply throw away the parts of $P_2$ that are
outside by deciding on which side of the intersection half-planes of $P_1$
these parts are. This is the basis for the Sutherland-Hodgman
algorithm~\cite{Sutherland}. For non-convex clip polygons one can still define
the inside of a polygonal curve by the winding number as done in the
Greiner-Hormann algorithm~\cite{Greiner}. However, we use a version of Vatti's
clipping algorithm~\cite{Vatti} which is slower in performance but does not
suffer from degeneracies as the Greiner-Hormann algorithm.

Now, we are equipped with a tool that finds the outline of the area of the
symmetric difference. Computing the measure of this area can then simply be
done by means of the divergence theorem, i.e., one needs to compute a
(discrete) boundary integral.

The computation of the gradient of $\mu(\mathcal{S}\sd\mathcal{T})$ with
respect to a variation of $\mathcal{S}$ is equivalent to computing the
gradient of $\mu((\mathcal{S}+u^h(\mathcal{S}))\sd\mathcal{T})$ with respect
to a small variation of the displacement field $u^h$ on the boundary
$\mathcal{S}$. Since $u^h$ is finite dimensional the gradient is a vector of
length $2K$ where $K$ is the number of boundary nodes as
in~(\ref{eq:5-7}). This is done by means of a finite difference scheme and
requires $2K$ applications of the clipping algorithm and $2K$ evaluations of
the area.

The number of applications of the clipping algorithm can be reduced to $K$ by
the following argument: By means of the divergence theorem we reduced the
computation of the area of the symmetric difference to the boundary of
$\mathcal{S}$. The gradient flow of the symmetric difference minimization
\begin{equation}
  \label{eq:5-8}
  \partial_t\mathcal{S} = -\nabla_{\mathcal{S}}\mu(\mathcal{S}\sd\mathcal{T})
\end{equation}
describes its evolution as a curve. The evolution of any curve along an
arbitrary velocity field $v$ is governed only by the contribution of $v$ that
points in shape's normal direction. The tangential part of $v$ just affects
the curve parametrization. But reparametrization does not affect the area of
the symmetric difference $\mu(\mathcal{S}\sd\mathcal{T})$. Since
$-\nabla_{\mathcal{S}}\mu(\mathcal{S}\sd\mathcal{T})$ is the direction of
steepest descent this rules out any contribution in tangential
directions. Hence, given the shape normals one can compute the gradient of the
symmetric difference by computing its directional derivative in normal
direction only. Since estimating shape normals is cheaper than polygon
clipping this reduces the computational cost for gradient calculation by
roughly $1/2$. This is an application of the Epstein-Gage lemma~\cite{Kimmel}
for curve evolution.

\section{Implementation and Experiments}\label{s-3}

The implementation was done on a standard desktop PC with a modern quad core
processor and 8GB of RAM using MATLAB. Looking at~(\ref{eq:5-7}) the reader
can see that there are three major bottleneck routines. First, one needs to
compute the matrix $S$ that describes the relation between boundary
displacements and boundary forces. For this we use our own implementation in
$C$.

The second bottleneck is the efficient computation of the symmetric
difference. For this we use, as said before, Vatti's algorithm. The C-library
GPC~\cite{GPC} implements this algorithm and is used in our
implementation. For two input polygons (clip and subject) the algorithm
generates a clipped polygon that outlines the intersection area of both inputs
or the area outlining the result of any other Boolean operation like the set
difference or the symmetric difference.

\begin{figure*}[t!]
  \centering
  \includegraphics[width=0.95\textwidth]{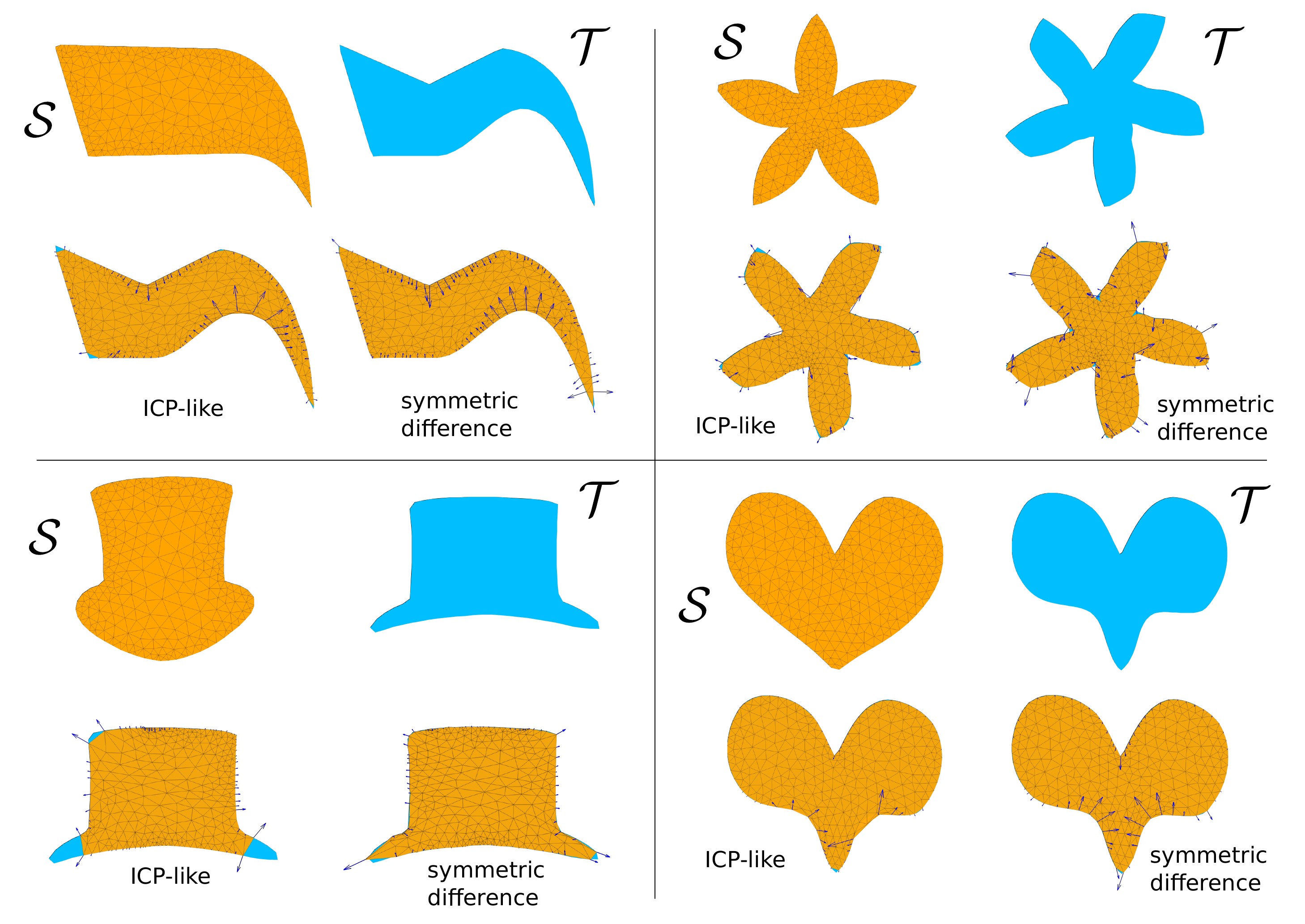}
  \caption[Symmetric difference method compared to the ICP-like method]{Four
    examples of matching in which the symmetric difference algorithm
    outperforms the ICP-like method. The latter produced flipped triangles in
    all examples as well as a bad alignment in case of the hat experiment
    (lower left). The elastic forces found by each algorithm are shown in blue
    together with an overlay of the matched shape and the target.}
  \label{c-4-fig-3}
\end{figure*}

\begin{figure}[h]
  \begin{center}
    \fbox{
      \includegraphics[width=0.45\textwidth]{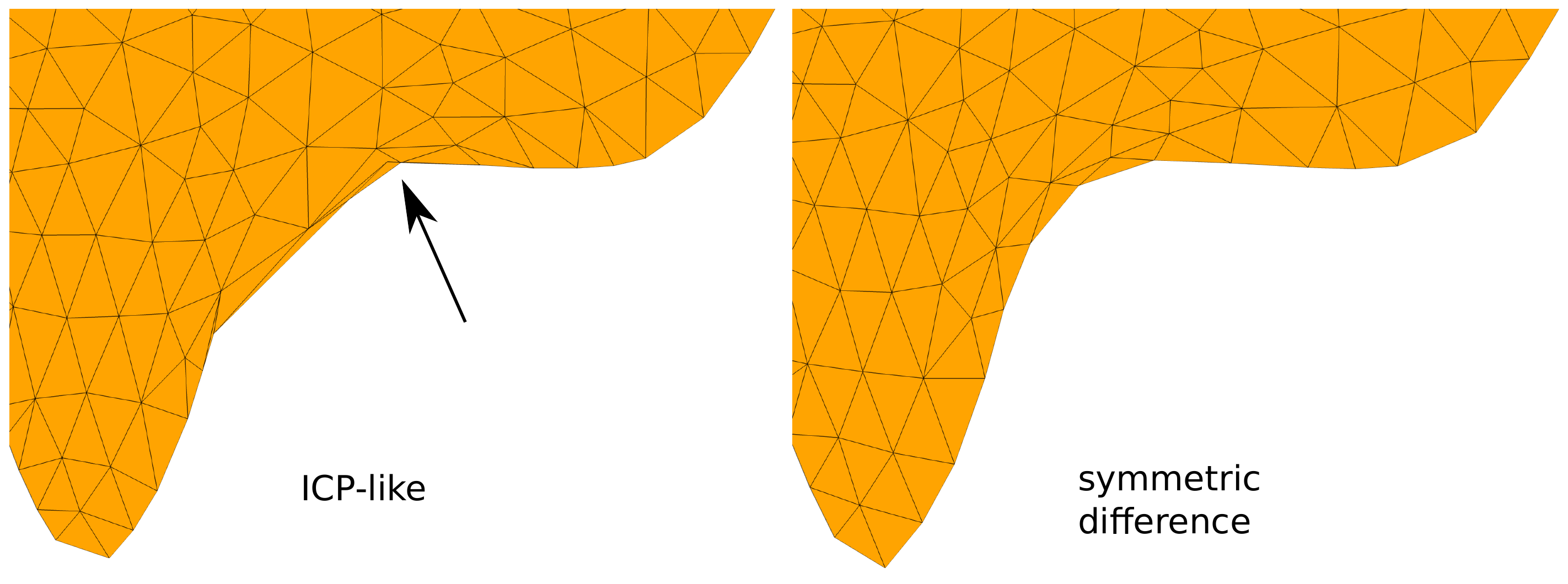}
    }
  \end{center}
  \caption[Close up of the heart experiment]{Close-up of the heart experiment
    shown in Figure~\ref{c-4-fig-3}. The ICP-like method produces flipped
    triangles in contrast to the symmetric difference method.}
  \label{c-4-fig-3a}
\end{figure}

\begin{figure*}[t!]
  \centering
  \includegraphics[width=0.95\textwidth]{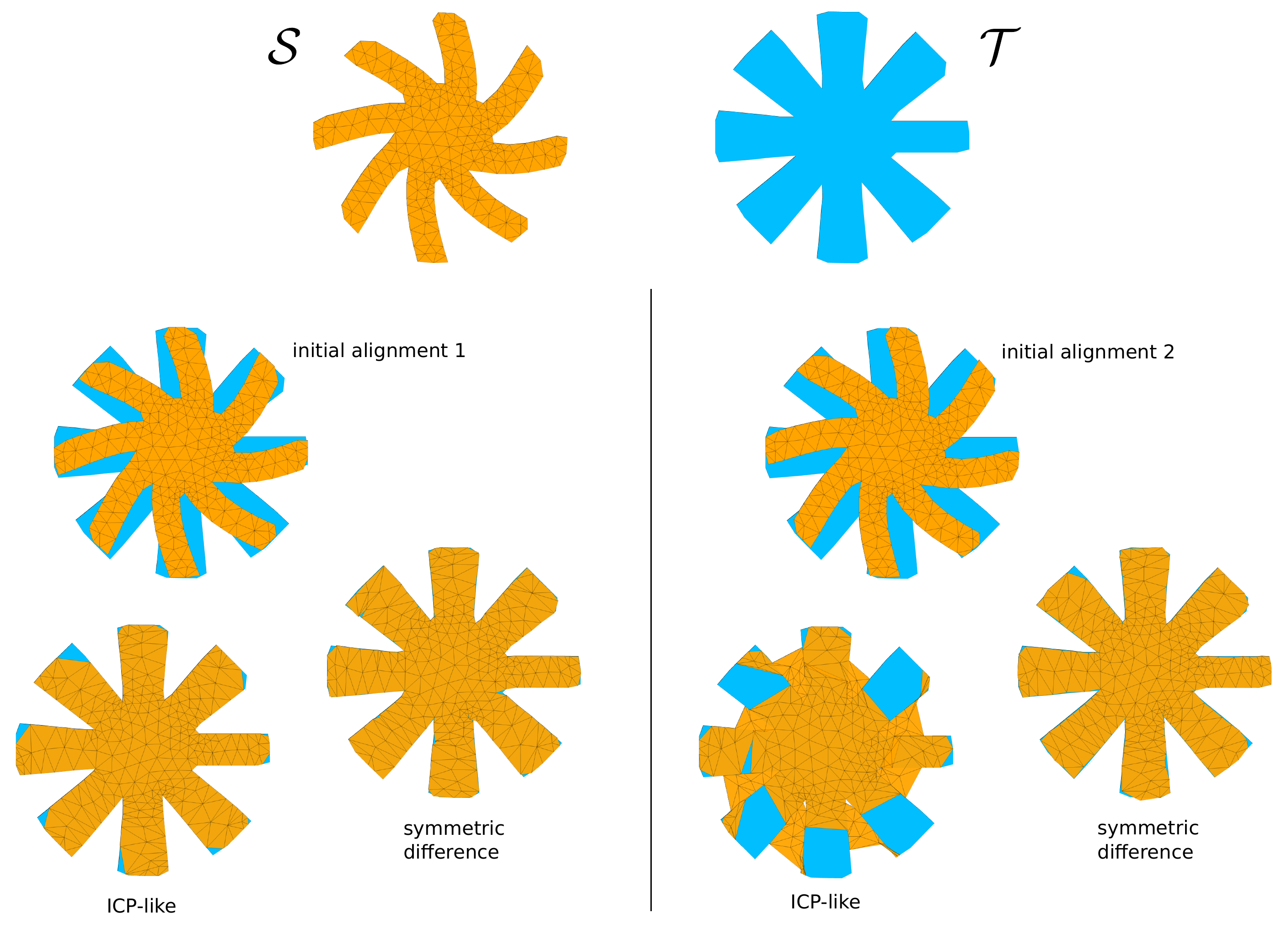}
  \caption[Robustness experiment under change of initial alignment]{Comparison
    of the effect of a slight change of the initial alignment of source
    $\mathcal{S}$ and target $\mathcal{T}$. The symmetric difference method
    still yields a good match while the ICP-like method fails for the second
    initial alignment. The initial alignments differ by a rotation of the
    source of 10 degree around its center of mass.}
  \label{c-4-fig-4}
\end{figure*}

The third bottleneck is the optimization procedure itself. For this note that
problem~(\ref{eq:5-7}) can be written in the form of a second order cone
problem
\begin{equation}
  \label{eq:5-9}
  \begin{split}
    \minimize_{f_i,d,e} & \quad \sum_{i=1}^K f_i + \alpha d + \beta e \\
    \text{subject to} & \quad \norm{S_iu_B}{2} \leq f_i \:,\quad i=1\dots K \:, \\
    & \quad \norm{ \left(
        \begin{array}{c}
          2 F(u^h)          
          \\ 
          d-1
        \end{array}
      \right) }{2} \leq d+1 \:, \\
    & \quad \norm{ \left(
        \begin{array}{c}
          2(u^h - u_0^h) \\ e-1
        \end{array}
      \right) }{2} \leq e+1 \\
  \end{split}
\end{equation}
where
\begin{equation}
  \label{eq:5-9a}
  \begin{split}
    F(u^h) & = \mu((\mathcal{S}+u_0^h(\mathcal{S}))\sd \mathcal{T}) \\
    & \quad + \nabla\mu((\mathcal{S}+u_0^h(\mathcal{S}))\sd
    \mathcal{T})(u^h-u_0^h) \:.
  \end{split}
\end{equation}
Among many good solvers for this type of problems we found that
MOSEK~\cite{MOSEK} is very efficient and used it through the YALMIP
interface~\cite{YALMIP}.

We conducted several experiments in order to show that our algorithm performs
well. The example shapes have been taken form~\cite{SebastianKleinKimia,Simon}
and the MPEG-7 dataset~\cite{MPEG7}. The first step was to extract the
(polygonal) boundary curve from the images that contain the shapes as a dense
ordered point set. Then we meshed the interior of the shapes to obtain a
Delaunay triangulation for the FEM. This is necessary to compute the stiffness
matrix of the relevant material. The triangulations contained between 200 to
700 triangles. As Lam\'e constants we chose $\mu=1$ and $\lambda=0$. This way
the material is allowed to stretch without shrinking in the lateral
direction~\cite{PeckarSchnorrRohrStiehl}.

For comparison we use the shape matching algorithm we introduced
in~\cite{Simon} as a baseline method. In the sequel we will refer to it as the
``ICP-like'' algorithm.  Recall that this algorithm is similar to ICP except
that we allow boundary correspondences to drift during the optimization.

The first experiment, see Figure~\ref{c-4-fig-1}, shows three simple shapes
that are matched to each other using our symmetric difference method. In the
figure we show, in addition to the shapes, the negative gradient of the area
of the symmetric difference in the initial position. This gradient can be
interpreted as a restoring or driving force that acts on the boundary of the
solid represented by $\mathcal{S}$. Note, that an initial overlap of the
source and the target is necessary for this gradient to contain information
about the desired final shape. The areas of the symmetric difference in each
iteration are shown in Figure~\ref{c-4-fig-1g}. One can see that on average
the area of symmetric difference decreases towards zero. We stopped the
algorithm at most 50 iterations or if the area of the symmetric difference
dropped below a certain threshold (1\% of the sum of the areas of the source
and the target). Also note that the deformations are quite large for a linear
material model.

\begin{figure*}[ht!]
  \centering
  \includegraphics[width=0.95\textwidth]{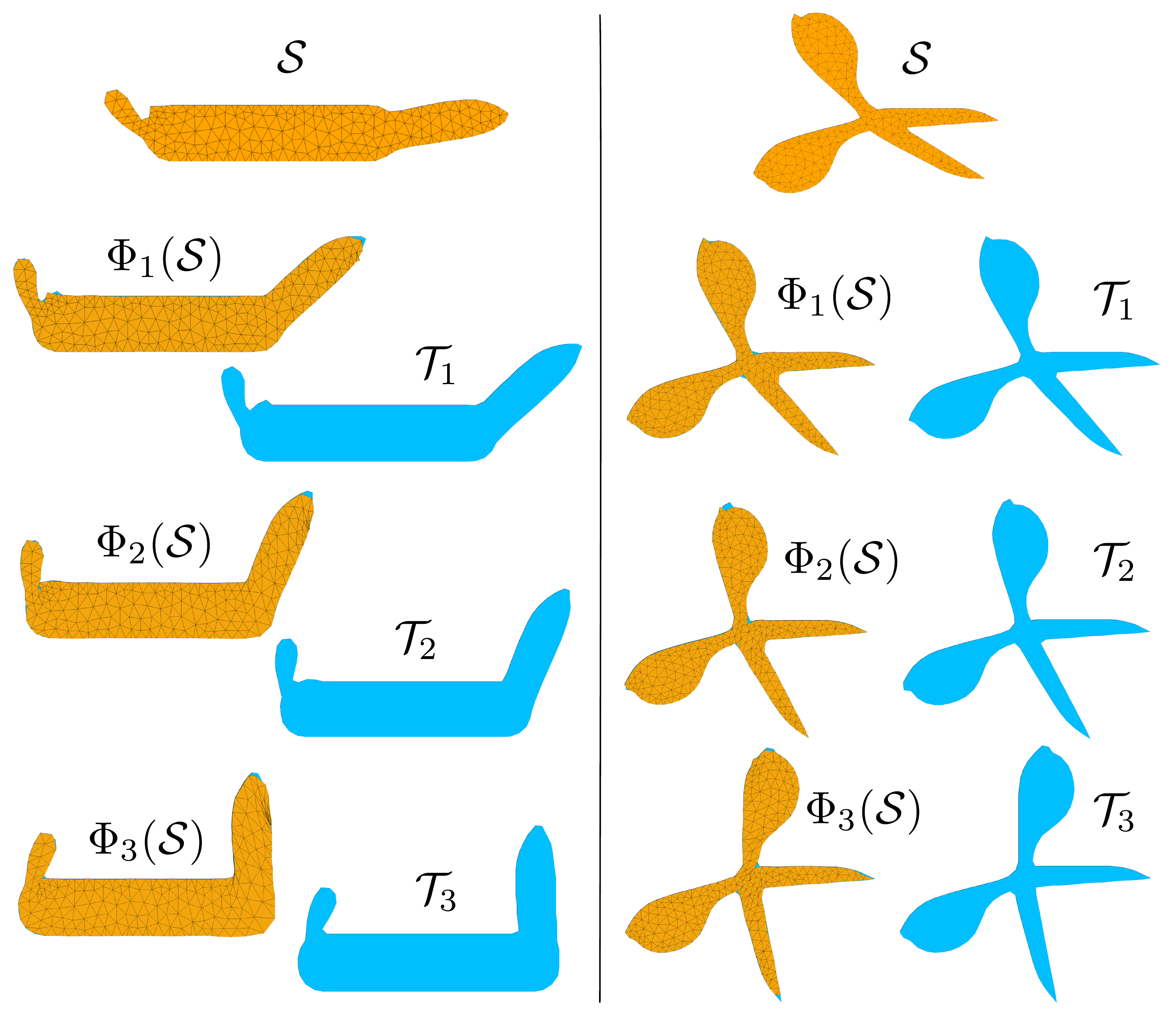}
  \caption[Gradually incresing deformations (Tool dataset)]{Two examples of
    gradually increasing deformations taken from the tool dataset described
    in~\cite{BBK,TOSCA}. The symmetric difference method produces good
    alignment but can suffer from large distortion and flipped triangles as
    the deformation becomes very large.}
  \label{c-4-fig-5}
\end{figure*}

In the second experiment, Figure~\ref{c-4-fig-2}, we match the shape of a
starfish to three different articulations. We compare the results of our
symmetric difference method to the results that we got by our ICP-like method
(using the same initial alignment) in terms of iteration numbers, norm of
forces and conformal distortion, see Table~\ref{c-4-t-1}. Although the results
appear to be visually similar the symmetric difference method generally
produced smaller elastic forces and conformal distortions (in maximum and
average) while slightly more iterations were needed.

\begin{table}[h]
  \begin{center}
    \begin{tabular}{ c | c | c | c }
      & $\mathcal{T}_1$ & $\mathcal{T}_2$ & $\mathcal{T}_3$ \\ \hline
      \textbf{\# Iterations}&  &  &  \\ \hline
      Symmetric difference & 19  & 21 & 19 \\ \hline
      ICP-like          & 14  & 15 & 14 \\
      \hline \hline
      \textbf{Norm of forces} &  &  &  \\ \hline
      Symmetric difference & 0.41  & 0.82 & 0.95 \\ \hline
      ICP-like          & 0.45  & 0.92 & 0.97 \\
      \hline \hline
      \textbf{Maximal CD} &  &  &  \\ \hline
      Symmetric difference & 2.15  & 12.46 & 1.89 \\ \hline
      ICP-like          & 4.04  & 98.1  & 2.92 \\
      \hline \hline
      \textbf{Mean CD} &  &  &  \\ \hline
      Symmetric difference & 1.14 & 1.18 & 1.18 \\ \hline
      ICP-like          & 1.15 & 1.35 & 1.24 \\
      \hline
    \end{tabular}
  \end{center}
  \caption{Comparison of the symmetric difference method to the ICP-like
    method in terms of number of iterations, norm of forces and conformal
    distortion (CD) in each triangle, see Figure~\ref{c-4-fig-2}.}
  \label{c-4-t-1}
\end{table}

Figure~\ref{c-4-fig-2a} shows the initial restoring force produced by the
symmetric difference method compared to the restoring forces that are
initially produced by the ICP-like method. The ICP-like forces are at many
places of the boundary unreasonable due to nearest neighbors that are far away
from the correct correspondences. The symmetric difference method produces
forces that tend to shrink the source $\mathcal{S}$ in regions that do not
overlap with the target $\mathcal{T}$ while it tends to expand the shape in
regions of overlap.


\begin{figure*}[t!]
  \centering
  \includegraphics[width=0.98\textwidth]{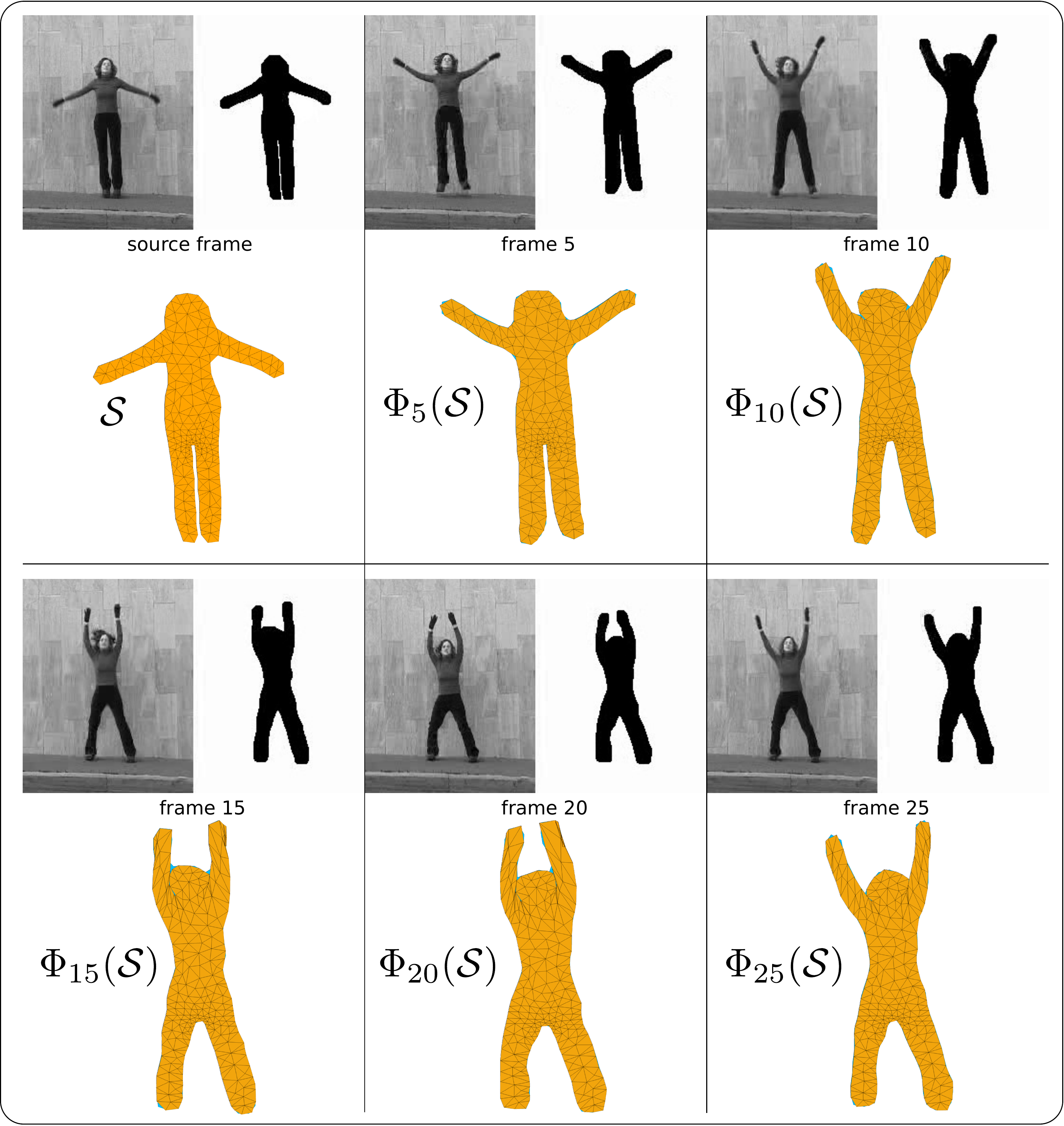}
  \caption[Jumping person experiment revisited]{We track the gradually
    increasing deformation of the silhouette of a jumping person taken from a
    video sequence that consists of 28 frames. The first frame serves as a
    source frame. We then match the source to the target silhouette in the
    $n$-th frame, starting from $n=2$. The resulting deformation was then used
    to initialize the method for matching the source to the target silhouette
    of the $(n+1)$-th frame. We show the result for the frame numbers 5, 10,
    15, 20, and 25. Note that although the deformation becomes quite large the
    method yields good alignment but can suffer frome flipped triangles and
    large conformal distortion.}
  \label{c-4-fig-6}
\end{figure*}


\begin{figure*}[t!]
  \centering
  \includegraphics[width=0.95\textwidth]{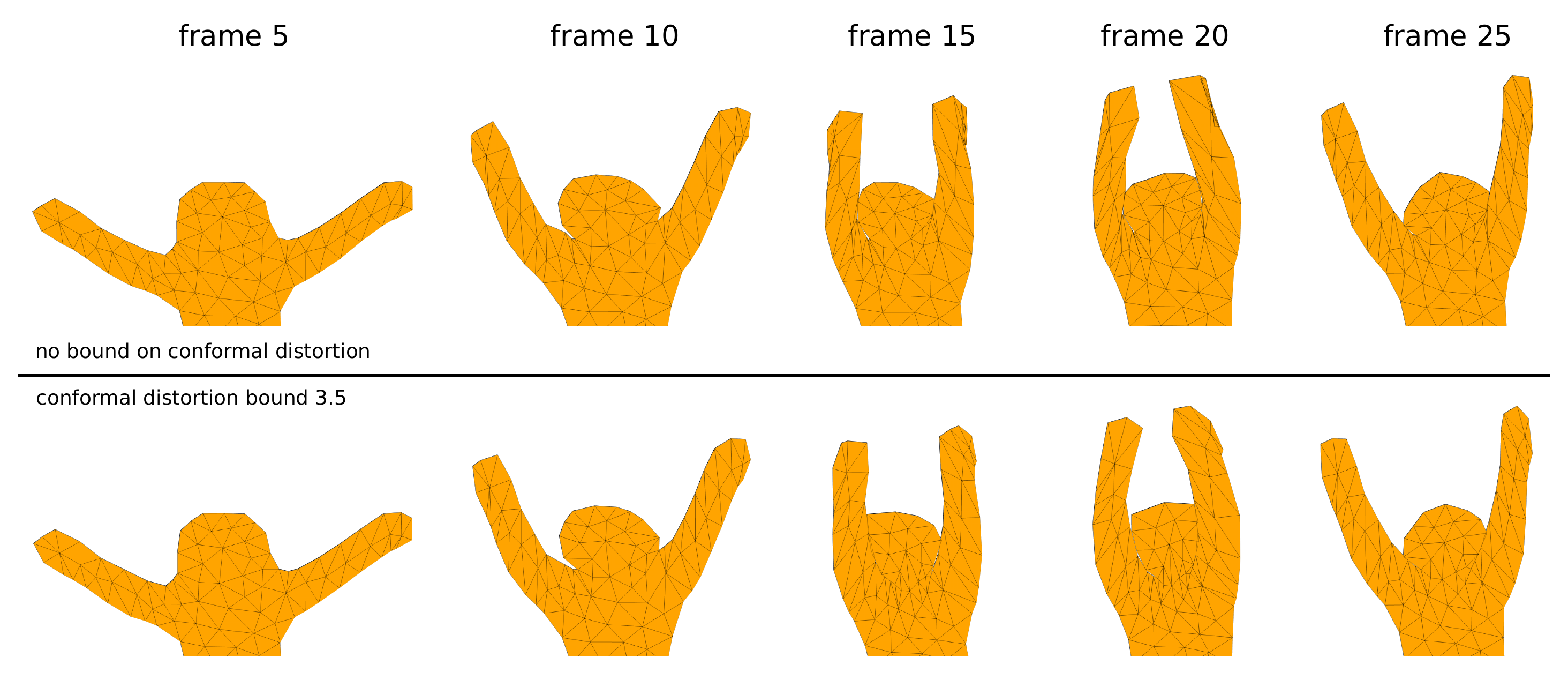}
  \caption[Jumping person: no distortion bound versus global distortion
  bound]{Close up of the experiment shown in Figure~\ref{c-4-fig-6} without
    distortion bound (upper row). We compare to it the visual effects of
    adding a global distortion bound of $3.5$ in the same experiment (lower
    row). The results in the lower row does not show flipped triangles and it
    is guaranteed that the maximal conformal distortion does not exceed the
    bound of $3.5$.}
  \label{c-4-fig-7}
\end{figure*}


\begin{figure*}[t!]
  \begin{center}
    \includegraphics[width=0.48\textwidth]{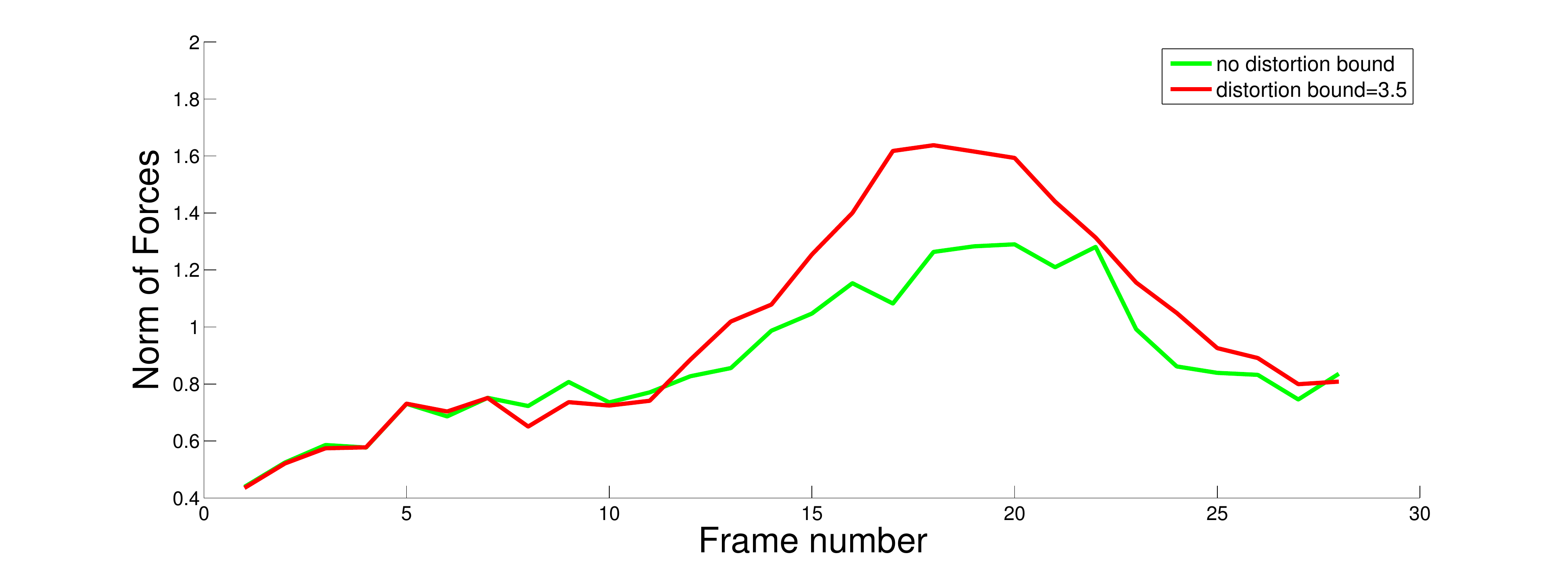}
    \includegraphics[width=0.48\textwidth]{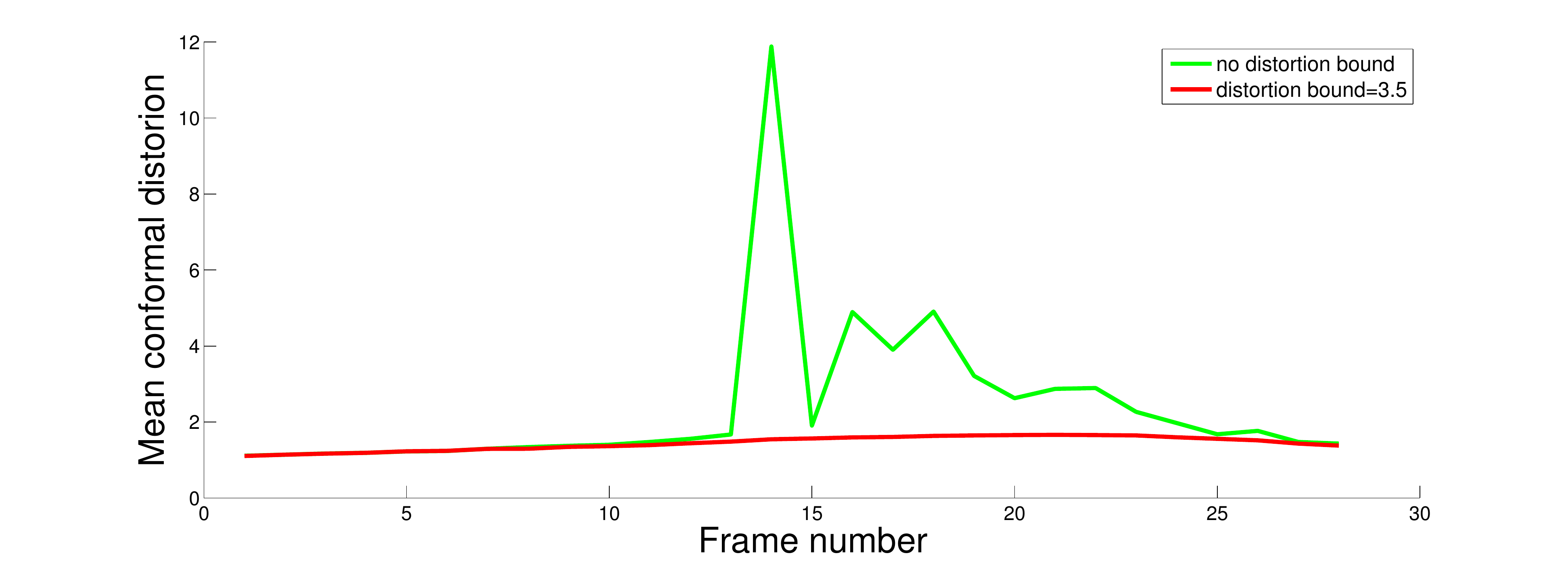}  
  \end{center}
  \caption[Jumping person: no distortion bound versus global distortion bound
  (graphs)]{The graphs compare the jumping person experiment shown in
    Figure~\ref{c-4-fig-6} to the same experiment with an additional global
    bound on the conformal distortion. Bounding the distortion leads to
    slightly higher forces (left) and lower distortion in average (right). The
    peak of the mean conformal distortion on the right-hand side (green)
    without distortion bound was reached in the frames when the deformation
    was most developed. For the red graphs we used a global distortion bound
    of $3.5$. See also Figure~\ref{c-4-fig-7} for a visual comparison.}
  \label{c-4-fig-6g}
\end{figure*}

Furthermore, we observed that in many scenarios the symmetric difference
method outperforms the ICP-like method. Figure~\ref{c-4-fig-3} shows four
examples in which the ICP-like methods produced not only higher conformal
distortion but also resulted in flipped triangles in contrast to the symmetric
difference method. A close-up view of one of the experiments is shown in
Figure~\ref{c-4-fig-3a}.

Both of these methods are local and hence sensitive to initial alignment of
source and target. As the shape matching problem is ill-posed the situation is
even worse since a slight change of the initial alignment can produce
completely different results of the methods. We observed in experiments that
the symmetric difference is often less sensitive to a slight change of the
initial alignment compared to the ICP-like method which relies on the
computation of nearest neighbors. This is demonstrated in
Figure~\ref{c-4-fig-4}.

Our method can be used to track gradually increasing
deformations. Figure~\ref{c-4-fig-5} shows two examples taken from the tool
dataset~\cite{BBK,TOSCA} and Figure~\ref{c-4-fig-6} shows the tracking of a
gradually increasing deformation taken from a video sequence used
in~\cite{Gorelik}. The video sequence, which depicts a jumping person,
contains the presegmented silhouettes of that person in each of the 28
frames. The silhouette of the first frame served as the source shape. We then
tracked shape change among the remaining frames using our symmetric difference
method. Once a good match was found between the source frame and the target
frame we used the result to initialize the method for matching the source to
the target shape of the succeeding frame. Note that in the later frames the
deformation becomes quite large. The same experiment was conducted
in~\cite{Simon} using the ICP-like method. The symmetric difference method
shows good alignment in the experiment shown in Figure~\ref{c-4-fig-5}
although some of the triangles suffer from flips as the deformation becomes
large. In the case of the video sequence, Figure~\ref{c-4-fig-6}, the
symmetric difference method also gives a good alignment (despite some flips)
among all frames of the sequence in contrast to the ICP-like method which
fails once the deformation becomes large.

So far we demonstrated the plausibility of the symmetric difference method and
showed that it can even outperform the ICP-like method. Both methods can
produce flipped triangles. It is possible to prevent the triangles from
flipping despite the fact that we reduced the elastic regularizer to the
boundary of the shape. To see this note that the nodal displacements of each
triangle in the source mesh describe an affine map for each triangle. The
distortion of each deformed triangle is controlled by the condition number of
the derivative of this affine map, i.e., by the ratio of its maximal to its
minimal singular value. The method introduced in~\cite{Lipman3} provides a
solution to bound the condition number, thus, bounding the maximal possible
(conformal) distortion and preventing flipped triangles. It is easy to
incorporate this method in our code since we simply need to know the
displacements of interior nodes, which can be found by a simple matrix
multiplication
\begin{equation}
  \label{eq:5-10}
  u_I = -A_{BI}A_{II}^{-1}u_B \:,
\end{equation}
see also equation~(\ref{eq:5-4a}). Hence, we have control over the interior by
means of the nodal boundary displacements $u_B$ and consequently over the
affine map of each triangle in a linear fashion. Ultimately, with regard to
the formulation of the optimization problem that we have to solve in each step
we simply have to add a second order cone constraint for each triangle. This
does not increase the optimization time dramatically. Figure~\ref{c-4-fig-7}
shows the visual improvement of controlling the conformal distortion. The
graphs shown in Figure~\ref{c-4-fig-6g} demonstrate that bounding the
conformal distortion results in slightly higher deformation forces and lower
conformal distortion. The higher deformation forces are due to the additional
constraints on the search space.

\begin{figure*}[t!]
  \centering
  \includegraphics[width=0.95\textwidth]{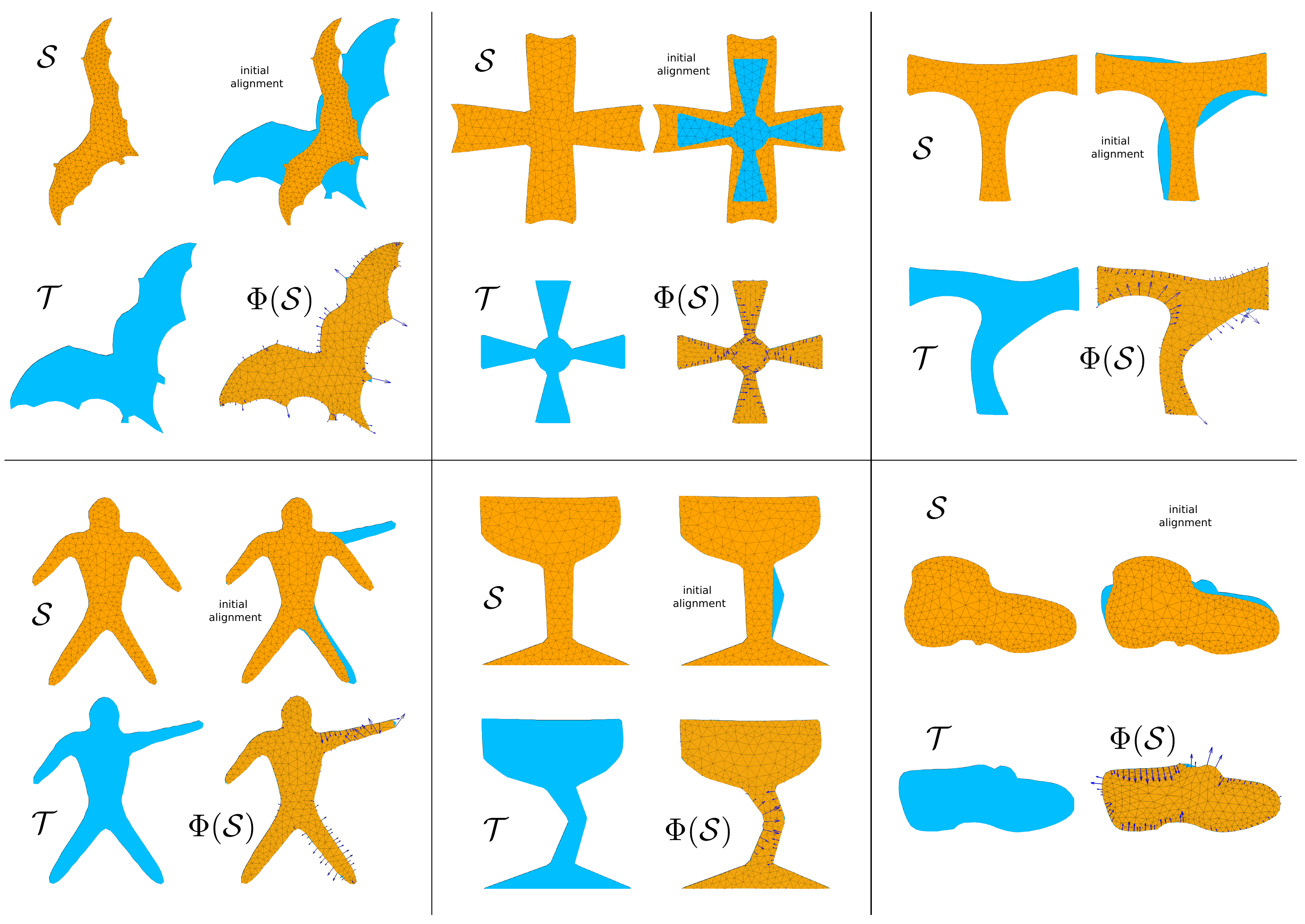}
  \caption[Additional experiments]{Additional experiments on the MPEG-7
    dataset. For each experiment we show the source shape $\mathcal{S}$ and
    the target shape $\mathcal{T}$ as well as their initial alignment. The
    matched shapes $\Phi(\mathcal{S})$ in each experiment are shown together
    with an overlap of the taget shape and the elastic forces found by our
    algorithm (blue arrows). No distortion bound was added.}
  \label{c-4-fig-8}
\end{figure*}

Figure~\ref{c-4-fig-8} shows additional experiments on shapes taken from the
MPEG-7 dataset. Here we did not add any distortion bound. The results are of
good quality and show intuitive elastic forces that cause the desired
deformation.


\section{Discussion}\label{s-4}

In this paper we introduced a novel method for shape matching of semantically
similar shapes in two dimensions. The technique is based on the observation
that shapes in many scenarios depict actual physical entities and hence shape
change can be measured by means of forces acting on them. We model the shapes
as plane elastic bodies and measure the severity of a deformation by means of
external elastic forces acting on their boundaries. Our method works on
triangular meshes but of course one can use any shape representation such as
polygons or point clouds to build such a mesh.

In order to describe the behavior of the shapes that we model as elastic
bodies we use the theory of linearized elasticity. The relevant equations are
the Navier-Lam\'e equations which constitute a linear system of partial
differential equations with additional boundary conditions. These can be
displacement conditions (Dirichlet type) and/or force conditions (Neumann
type) and they are usually unknown in the context of shape matching. We show
how to map displacement boundary conditions linearly to boundary forces in a
FEM framework. This enables us to seek boundary displacements that supplement
the Navier-Lam\'e equations such that their physical cause, i.e., the boundary
force, is of a simple nature. As a measure of simplicity, we believe, that the
sparsity of forces is a good prior since for a given object such forces can
describe different articulations of that object by mostly acting on the
articulated parts.

As a (dis-)similarity measure we use the area of mutual non-overlap of the
source and the target shape, i.e., the area of the symmetric difference of the
two shapes. This is a global measure and its minima under deformation of the
source mean a maximum overlap of the shapes. This is clearly in contrast to
classical ICP since a minimal mean (Euclidean) distance (under deformation of
the source) can be obtained by mapping the source to a single point (without
regularization).

Our algorithm is designed so that it simultaneously seeks low sparse forces
and a good alignment, measured by the area of the symmetric difference. We
show a heuristic way to approximate the nonlinear optimization problem by a
sequence of (convex) second order cone problems that only involve the boundary
of the mesh. This reduces the number of variables of the problem. We apply the
algorithm to a number of examples taken from various datasets. In addition, we
provide a comparison to our ICP-like method introduced in~\cite{Simon} and
show that the new symmetric difference algorithm can outperform the ICP-like
method.

Although the Navier-Lam\'e equations are only valid in the narrow regime of
small displacements and strain we show that the symmetric difference method
can find remarkably large deformations in which the ICP-like method
failed. Since we use gradient information of the symmetric difference the
method is, similarly to the ICP-like method, a local method that is sensitive
to the initial alignment of the shapes.

Despite good alignment results the symmetric difference method can
unfortunately suffer from large conformal distortions of the triangular mesh
and elements can even flip. We therefore suggest a way to deal with this
problem by incorporating into our optimization a method that is guaranteed to
have no flipped triangles and does not exceed a distortion bound that is
pre-selected by the user. Incorporating this method, which was introduced
in~\cite{Lipman3}, does neither change the type of the optimization problem
nor its dimensionality. One simply has to add a number of second order cone
constraints. The distortion bound should not be chosen too low since a low
bound might exclude a good solution.

Unfortunately this algorithm does not generalize to three dimensions because
the clipping algorithm does not generalize. Also, a generalization to
nonlinear material models, i.e., to rotation invariant deformation cost is
desirable in order to extend the applicability of the method. This is left for
future research.

\vspace{0.5cm}

\textbf{Acknowledgments.}  This research was supported by the Israel Science
Foundation, Grant No. 1265/14. The vision group at the Weizmann Institute is
supported in part by the Moross Laboratory for Vision Research and Robotics.

\end{document}